\documentclass[runningheads]{llncs}
\usepackage[table,xcdraw]{xcolor}

\usepackage[T1]{fontenc}
\usepackage{graphicx}
\usepackage{soul}
\usepackage{color}
\usepackage{url}
\usepackage[hidelinks]{hyperref}
\usepackage[small]{caption}
\usepackage{amsmath, amsfonts, amssymb}
\usepackage{booktabs}
\usepackage{algorithm}
\usepackage{algorithmic}
\usepackage[switch]{lineno}
\usepackage{subcaption}
\usepackage{multirow}

\usepackage{latexsym}

\begin{document}
\title{Revisiting the Disequilibrium Issues \\ in  Tackling Heart Disease Classification Tasks}
\titlerunning{Disequilibrium Issues in Heart Disease Classification}
%
\author{Thao Hoang\inst{1} \and
Linh Nguyen\inst{1} \and
Khoi Do\inst{1,}\thanks{Project Leader} \and
Duong Nguyen\inst{2} \and
Viet Dung Nguyen\inst{1}}

\authorrunning{Thao Hoang et al.}
%
\institute{Hanoi University of Science and Technology, Pusan National University\\
\email{\{hoangminhthao1111, nguyenphuonglinh2611, khoido8899, mduongbkhn, nvdung.bme, danghoanglan271\}@gmail.com}}

\maketitle              
\begin{abstract}
In the field of heart disease classification, two primary obstacles arise. Firstly, existing Electrocardiogram (ECG) datasets consistently demonstrate imbalances and biases across various modalities. Secondly, these time-series data consist of diverse lead signals, causing Convolutional Neural Networks (CNNs) to become overfitting to the one with higher power, hence diminishing the performance of the Deep Learning (DL) process. In addition, when facing an imbalanced dataset, performance from such high-dimensional data may be susceptible to overfitting. Current efforts predominantly focus on enhancing DL models by designing novel architectures, despite these evident challenges, seemingly overlooking the core issues, therefore hindering advancements in heart disease classification. To address these obstacles, our proposed approach introduces two straightforward and direct methods to enhance the classification tasks. To address the high dimensionality issue, we employ a Channel-wise Magnitude Equalizer (CME) on signal-encoded images. This approach reduces redundancy in the feature data range, highlighting changes in the dataset. Simultaneously, to counteract data imbalance, we propose the Inverted Weight Logarithmic Loss (IWL) to alleviate imbalances among the data. When applying IWL loss, the accuracy of state-of-the-art models (SOTA) increases up to $5\%$ in the CPSC2018 dataset. CME in combination with IWL also surpasses the classification results of other baseline models from $5\%$ to $10\%$. 

\end{abstract}
\section{Introduction}
In recent times, there has been a remarkable rise in the number of studies on applications of Artificial Intelligence (AI) in the healthcare field
\cite{ijcai2021p495,ijcai2021p219}. There are many benefits associated with this tendency, such as in the areas of disease classification \cite{8953911,CASENEUVE2021658,8099852}, drug discovery \cite{10058590,9715218} and tumor segmentation \cite{ex,9547084}.

Subclinical diagnoses for heart disease classification mostly depend on the Electrocardiogram (ECG) signal's evolving patterns over time.
12 leads are used in the acquiring process; each lead depicts a distinct view of the ECG signal.
Although there exists a noticeable imbalance in magnitude amongst channels, some techniques aim to enhance outcomes by reconfiguring model \cite{1DNN,7vision,deform}, transfer learning \cite{transfer2,Weimann2021,transfer} and employing feature extraction \cite{Soaro}. Nevertheless, when utilized on frequency-varied biomedical datasets, particularly ECG signals, these methods exhibit a disregard for the variance in power distribution for every ECG lead and result in an increase in model size. In low-magnitude signal channels, the model cannot extract the essential information effectively, even with feature extraction methods. In addition, not having been thoroughly investigated, the possible risks of the channel-wise imbalance would result in the AI model receiving insufficient data. 

Another significant concern is that data collected in clinical settings often face challenges due to a long-tailed distribution of classes.
Therefore, several methods have been developed: oversampling data \cite{smote,oversamp}, undersampling data \cite{undersamp}, and data augmentation \cite{augment1}. All mentioned approaches aims to equilibrate the distribution of classes.
However, while oversampling may aid in the reduction of imbalances, it carries the danger of causing the model to become overfit to long-tailed classes. Undersampling, on the other hand, risks losing vital information, potentially limiting the model's ability to generalize to new data. Meanwhile, data augmentation can inadvertently augment the class imbalances by generating more samples from undesired classes \cite{survey}.

This paper aims to address the previously mentioned issues with the ECG dataset.
 First, a thorough empirical investigation into each channel's variation is carried out to obtain a greater understanding of the differences between the signal leads.
 Taking note of each lead's diminishing distribution and signal power, we propose a Channel-wise Magnitude Equalizer (CME) (see Fig. \ref{fig:method-overview}) method. In order to preserve the frequency characteristics while also improving the information of the fading signal, we normalized the data and used time-frequency interpolation. 
 Through this process, the power of acquired data is equalized with that of raw data. Second, we propose the Inverted Weight Logarithmic (IWL) loss function to tackle class imbalance issue without complicating AI models. In summary, our main contributions to this paper are as follows:
\begin{itemize}
\item A thorough empirical study is undertaken on variations in ECG leads in order to gain a profound insight into the fundamental properties of the signal.
\item Propose CME method to address power imbalances and distribution variation in fading signals, and improve the potential information contained while preserving the signal frequency characteristics. 
\item Propose IWL loss function to tackle the data distribution imbalance problem.
\item Experiment to verify the efficacy of our suggested method by equitably contrasting it with multiple baselines. Besides, an ablation test to evaluate the method's adaptability is also taken into account. 
\end{itemize}
\section{Methodology}
\begin{figure}[!ht]
    \centering
    \includegraphics[width = 1.043\linewidth]{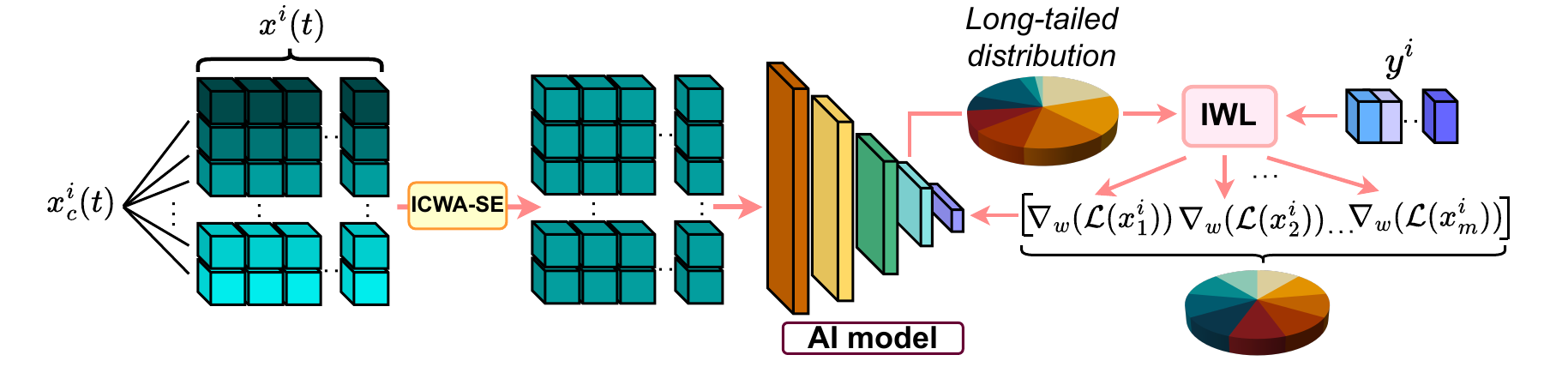}
    \caption{Methodology Overview: The input channel-wise ECG signal first goes through the CME process. Here, we normalize each channel-wise signal and assess its contribution (illustrated by different tones of green color) to the loss update by using an Attention-like layer. The $C$ channels are then squeezed and excited based on their relative importance, followed by an interpolation. The obtained images are then fed to the AI model. The original dataset has a long-tailed distribution. After each training epoch, the model gives prediction vectors, and together with true labels $y^i$, they go into the IWL loss function. Here, IWL loss performs gradient balancing and updates to the AI model to strengthen the weight from long-tailed classes and weaken the data from high-frequency classes.} 
    \label{fig:method-overview}
\end{figure}

The flow of the proposed methodology is presented in Fig. \ref{fig:method-overview}.

\subsection{Preliminary and Notations}
Given a dataset containing $N$ samples represented as $\{\textbf{x}^i, \textbf{y}^i\}_i^N$ with the total of $M\sim p(M)$ classes, each sample belonging to class $m$ is represented as $(\textbf{x}_m^i, \textbf{y}_m^i)$, where $m\in\left[1, \dots, M\right]$. We defined the model's weight as $\theta$, and the purpose is to minimize the loss function $\mathcal{L}$. For instance, 
\begin{equation}
    \label{eq:overall}
    \underset{\theta}{\min}~\mathcal{L}(\theta) \triangleq \mathbb{E}_{\textbf{x}^i,\textbf{y}^i\sim P(\textbf{x},\textbf{y})} [\ell (\textbf{x}^i,\textbf{y}^i\vert \theta)].
\end{equation}
Noted that the main consideration can be perceived as a classification task, we reshape the problem as presented in (\ref{eq:overall-cls}):
\begin{equation}
    \label{eq:overall-cls}
     \underset{\theta}{\min}~\mathcal{L}(\theta) = \frac{1}{N}\frac{1}{M}\sum_{i=1}^{N}\sum_{m=1}^M[\ell (\textbf{x}^m_i,\textbf{y}^m_i\vert \theta)],
\end{equation}
with $\mathcal{L}(\theta)$ is the cross entropy function, and can be represented as follows:
\begin{equation}
    \label{eq:ce}
    \mathcal{L}(\textbf{x}^i, \textbf{y}^i) = -\sum_{m=0}^{M-1}\log\left(\frac{\exp\{f_\theta(\textbf{x}_m^i)\}}{\sum_{m'=0}^{M-1}\exp\{f_\theta(\textbf{x}_{m'}^i)\}}\right)\textbf{y}^i_m
\end{equation}
with $f_\theta(\cdot)$ is the classification task of the given AI model.
\subsection{Empirical Study}\label{sec:empirical-analysis}
\begin{figure}[!ht]
    \centering
    \begin{subfigure}{0.35\columnwidth}
        \centering
        \includegraphics[width = \textwidth]{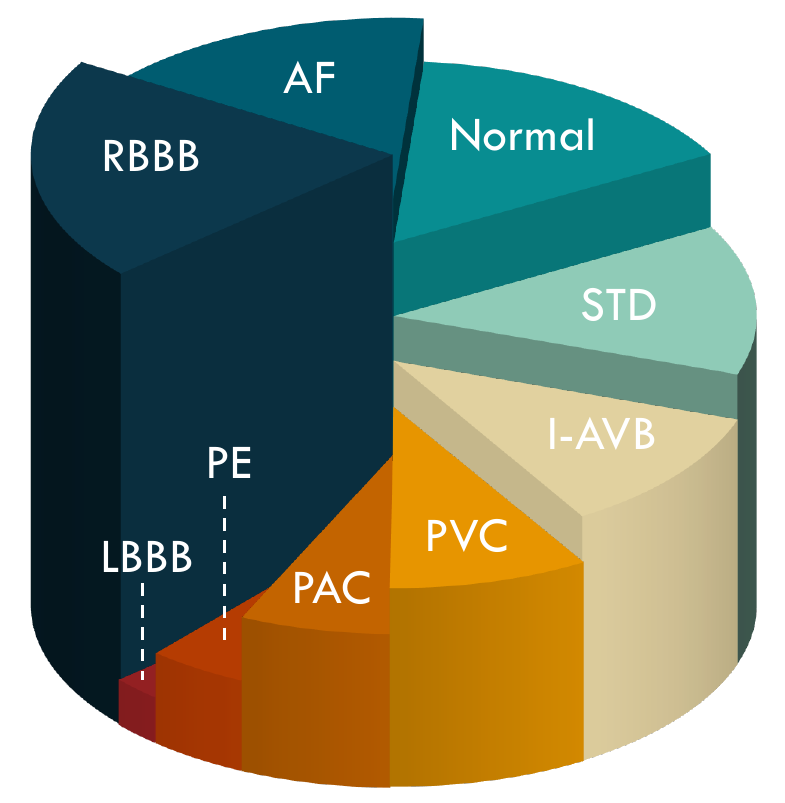}
        \caption{Class Frequency Imbalance}
        \label{fig:ecg_sig}
    \end{subfigure}
    \begin{subfigure}{0.45\columnwidth}
        \centering
        \includegraphics[width = \textwidth]{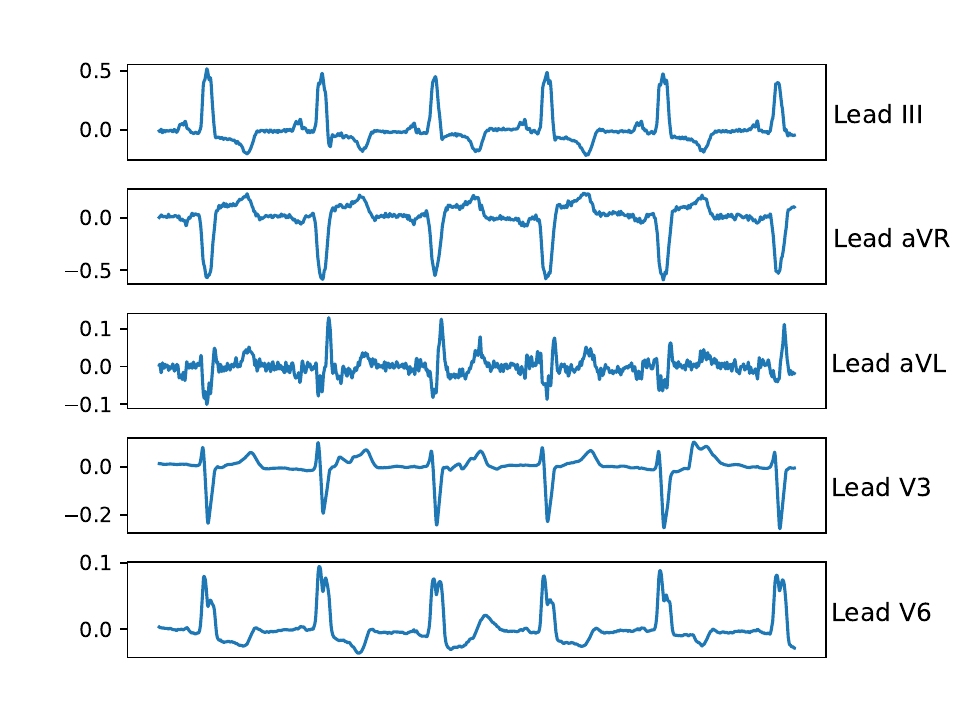}
        \caption{Channel Magnitude Imbalance}
        \label{fig:cls_freq}
    \end{subfigure}
    \caption{Dataset illustration: a) The sampling distribution of 9 classes (total of 6400 single label samples): Right bundle branch block (RBBB), Atrial fibrillation (AF), Normal, ST-segment depression (STD), First-degree atrioventricular block (I-AVB), Premature ventricular contraction (PVC), Premature atrial contraction (PAC), ST-segment elevated (STE) and Left bundle branch block (LBBB). b) The ECG waveforms (Left bundle branch block disease, from A0011). There are 5 leads being illustrated.}
\end{figure}

To achieve a comprehensive illustration of the uncovered problems, we take a deep analysis of the data (refer to Fig. \ref{fig:ecg_sig}), where the input data $\textbf{x}^i \in \mathbb{R}^{C \times L}$ is considered as a composition of multiple stochastic random processes $x^i_c(t)$, where $C$, $L$ is the number of channels and channel signal length. Easily seen from Fig. \ref{fig:ecg_sig}, the primary issue while learning patterns from the dataset is that the maximum level of each channel varies significantly. Consequently, a magnitude gap exists between learning gradients due to the fact that the input channels with higher magnitudes will contribute to the higher loss value. For instance: 
\begin{align}
    \label{eq:ce2}
    \mathcal{L}(\textbf{x}^i, \textbf{y}^i) 
    &= -\sum_{i=0}^{N-1}\sum_{m=0}^{M-1}\sum_{c=0}^{C-1}\log\left(\frac{\exp\{f_\theta(\textbf{x}^i_{m,c})\}}{\sum_{m'=0}^{M-1}\exp\{f_\theta(\textbf{x}^i_{m',c})\}}\right)\textbf{y}^i_{m,c} \notag \\ 
    &= -\sum_{i=0}^{N-1}\sum_{m=0}^{M-1}\sum_{c=0}^{C-1}\log\Big(\frac{\exp\{f_\theta(\beta_{m,c} \Bar{\textbf{x}}^i_{m,c})\}}{\sum_{m'=0}^{M-1}\exp\{f_\theta(\beta_{m,c} \Bar{\textbf{x}}^i_{m',c})\}}\Big)\textbf{y}^i_{m,c} \notag \\
    &= -\sum_{i=0}^{N-1}\sum_{m=0}^{M-1}\sum_{c=0}^{C-1}\beta_{m,c} \log\Big(\frac{\exp\{f_\theta(\Bar{\textbf{x}}^i_{m,c})\}}{\sum_{m'=0}^{M-1}\exp\{f_\theta(\beta_{m,c} \Bar{\textbf{x}}^i_{m',c})\}}\Big)\textbf{y}^i_{m,c},
\end{align}
where $\Bar{\textbf{x}}^i_{m,c}$, $\beta_{m,c}$ represents the normalized signal of data $\textbf{x}^i$ and the magnitude scale according to class $m$ and channel $c$, respectively. As easily seen from Equation \eqref{eq:ce2}, the loss will vary significantly according to different leads' power levels. Moreover, due to the nature of the ECG sampling process, there occasionally occurs an unstable power level in one lead. Thus, conventional normalization is unable to mitigate the inconsistent and imbalanced issues in ECG classification.

\subsection{Squeeze and Excitation with Channel-wise Magnitude Equalizer}
\label{sec:SE-ICWA-SE}
From the comment in Section~\ref{sec:empirical-analysis}, we proposed a simple method for our model architecture's first layer (Fig. 1). 
 We used an Attention-like layer \cite{2017-DL-Attention} to quantify the contribution of each channel to the loss update. The SE approach from \cite{2018-DL-SENET} is then used to scale each channel based on its relative relevance.

Instead of the traditional structure, we applied an inverted version, CME. Our goal is to excite the channels with lower loss update contribution and squeeze the channels with higher contribution, i.e. as follows:
\begin{align}
    k^i_c = F_{\textrm{CME}}(z^i_{\cdot, c}) = \frac{e^{-\Vert\textbf{x}^i_{\cdot, c}\Vert}}{\sum_{c'=0}^{C-1} e^{-\Vert\textbf{x}^i_{\cdot, c}\Vert}},
\end{align}
where $k^i_c$ represents the squeeze and excitation factor according to channel $c$. Applying all channels through the input data $\textbf{x}^i_c$, we have the set of factor $k = \{k^i_c \vert ~c=1,2,\ldots,C\}$. Subsequently, we leverage the SE factor to scale the channel-wise representation $\textbf{z}^i_{\textrm{SE}}$ as follows: 
\begin{align}
    \textbf{z}^i_{\textrm{SE}} = F_{\textrm{scale}}(\textbf{x}^i, k),
\end{align}
where $F_{\textrm{scale}}(\cdot, \cdot)$ is the Hadamard product function, represented as follows: 
\begin{align}
    F_{\textrm{scale}}(x, k) = x \cdot k.
\end{align}
By doing this step, the channel-wise signal, i.e., leads, are automatically squeezed and excited according to the instantaneous signal. To this end, these scaled signals are interpolated to not just enhance the quality of information but also keep the frequency properties unchanged. Note that, this CME block is a hyperparameter-free approach that equalizes the gradient magnitude across tasks to encourage learning low power $\mathcal{P}(x^i_c(t)) = \frac{1}{T}\sum_{t = 0}^{T - 1}\|x^i_c(t)\|$ signal and difference in signal distribution among channel ($p(x^i_h) \neq p(x^i_k),~\forall h \neq k$). 
\subsection{Equalized Scalarization via IWL Loss}
As mentioned in the previous sections, data imbalance is a critical issue in learning biomedical data in general and ECG classification in particular. To tackle the problem efficiently, we proposed a new loss function which is IWL (Inverted Weight Logarithmic) loss to deal with a long-tailed dataset:
\begin{align}
    \label{eq:iwl}
    \mathcal{L} &=  -\dfrac{1}{N} \sum^{N-1}_{i=0}\sum^{M-1}_{m=0}\left(log\left(\dfrac{10}{p(\hat{x}^i_m)+\varepsilon}\right)\right)^\beta log(p(\hat{x}^i_m))y^i_m,
\end{align}
where $p(\hat{x}^i_m)\in\mathbb{R} = \exp\{\hat{x}^i_m\}/\sum_{m'=0}^{M-1}\exp\{{\hat{x}^{i}_{m'}}\}$ is the normalized prediction vector and $\hat{x}^{i}_{m} = f_\mathcal{W}(x^{i}_{m})$, respectively. $\varepsilon$ is an infinitesimal and $\beta$ is temperature parameter. IWL can put a higher weight onto the low confidence logits while keeping a small weight for well-classified ones. This feature makes the model focus more on classifying long-tailed rare categories and avoid forgetting high confidence classes, compared to Focal\cite{focal} or CBFocal\cite{clsballoss}. 
\section{Experimental Evaluation}
\subsection{Experimental Settings}\label{sec_3.1}
\textbf{Problems.} We conduct three large experiments (see Table \ref{table:compare}). First, we reimplement state-of-the-art models and techniques that use the same dataset as ours, which include: Auto Encoder \cite{Soaro}, 1D DNN \cite{1DNN}, BiRNN \cite{birnn}, BiLSTM \cite{bi_lstm}, Transformer \cite{review}, ResNet \cite{res}, and augmentation. Besides, we replace the loss function used in those models in the first experiment with our IWL loss and compare the results obtained from two loss functions. Second, we apply different loss functions to our ICWA-SE trained by EfficientNetB0, which are: CB Focal \cite{clsballoss}, Class-balanced \cite{clsballoss}, Focal \cite{focal} and LDAM \cite{labloss} loss which are SOTA loss functions that popularly used for imbalanced dataset. In each loss, we utilize the best hyperparameters that are justified by their authors. Third, we conduct some model architectures (RegNet \cite{reg}, ResNeXt \cite{next}, EfficientNetV2 \cite{v2}) combined with our proposed ICWA-SE and IWL loss function. 

For ours IWL loss, we provide a configuration with ten different values of $\beta \in $ \{0.1, 0.3, 0.5, 0.7, 0.9, 1, 2, 3, 4, 5\} to find the optimal one (see Section \ref{secabl1}). \\ 
\textbf{Datasets.} To evaluate the performance of the proposed methods, we make use of the CPSC2018 dataset \cite{liu2018open}, the dataset consists of 6877 12-lead ECG recordings sampled at 500Hz, each recording has a length from 6s to 60s. However, as there exist several samples with multiple labels, in our experiments, only 6400 samples with single labels will be used to avoid conflicted patterns between multi-label samples and single-label samples. We then split the obtained dataset into the training set, and test set by the ratio $[0.9, 0.1]$, respectively. For the ICWA-SE method, after cutting off 500 data points at the beginning (to prevent the artifact from the first second of the acquisition process), only 2500 data points from each sample were used for implementation. Since the shortest samples from the dataset have a length of 3000 data points (equivalent to 6s at 500Hz sampling rate), we only use the first 3000 data points of all datasets to avoid differences in the size of the input. 

To deeply evaluate our proposed method with imbalanced data, we shuffle the dataset using the following equation $N = max(x_m^i) * \alpha^{(m/ (M - 1)}$ with $\alpha \in \{0.05, 0.01\}$, note $\alpha$ as a parameter represent the imbalance dataset, the smaller the $\alpha$ value, the more imbalance the data is, $\alpha=0.05$ claims the distribution of the original dataset. \\ 
\textbf{Experiment setup and Evaluation metrics.} We conduct experiments on 2D CNN models that are mentioned above. The data was trained over 150 epochs with learning rate equals 0.001 and Adam optimizer. The batch size is set to 64. Two metrics accuracy (\%) and F1-score (\%) are used for performance evaluation.

\subsection{Overall Performance}\label{per}

Table \ref{table:compare} shows the results of all experiments that we mentioned in Section \ref{sec_3.1}. Overall, the highest accuracy and F1-score are achieved with our method, which outperforms others in both $\alpha$ values. In most experiments, it's observed that the outcomes from the original dataset surpass those obtained with smaller $\alpha$ values. However, the performance of the ICWA-SE model with the IWL loss demonstrates a contrary trend, where the superior outcome is achieved with the dataset possessing a smaller imbalanced factor ($\alpha = 0.01$). This suggests the capability of our proposed method to handle highly imbalanced data effectively.\\
\textbf{Comparison to State-of-the-art models.}
We bring into comparison four SOTA models using two loss functions, Cross-Entropy loss and our IWL loss. From Table \ref{table:compare}, it is noticable that the accuracies of most SOTA models with IWL loss are higher compared to those without its adoption (from 0.5\% to 5\%), which can prove the effectiveness of our IWL loss function.\\
\textbf{Comparison to State-of-the-art loss functions.}
To examine the combination of the IWL loss when being applied on CME method, five SOTA loss functions have been used for comparison (trained with EfficientNetB0 model). The obtained results with four loss functions are higher than all the reimplemented SOTA models (except for CB Focal loss), proving the efficiency of our CME method. It is also noted that our IWL loss function give the best results on the proposed CME method compare with the other five loss functions used, which can prove the effectiveness combination of CME method and IWL loss.

\begin{table}[!ht]
\centering
\caption{Comparison of our proposed model with other models over two alpha values. Each method is tested with two scenarios including $\alpha = 0.05$, which is the original dataset, and $\alpha = 0.01$ whose imbalance among classes is stronger. The results that obtained using our CME and IWL loss is bolded to denote that our methods outperform others in experiment.}
\resizebox{\textwidth}{!}{%
\begin{tabular}{|l|l|cc|cc|}
\hline
 &
   &
  \multicolumn{2}{c|}{$\alpha = 0.05$ (Original)} &
  \multicolumn{2}{c|}{$\alpha = 0.01$} \\ \cline{3-6} 
\multirow{-2}{*}{\textbf{Method}} &
  \multirow{-2}{*}{\textbf{Loss}} &
  \multicolumn{1}{c|}{\; Accuracy \;} &
  \; F1-score \;&
  \multicolumn{1}{c|}{\; Accuracy \;} &
  \; F1-score \;\\ \hline
(1D) Auto Encoder \cite{Soaro}&   & 
 $67.8 \pm 1.2$       & $58.0 \pm 1.6$        & $67.9 \pm 1.1$     & $59.5 \pm 1.3$\\
(1D) DNN \cite{1DNN}&   & $73.0 \pm 1.5$        & $48.7 \pm 1.6$                      & $71.7 \pm 1.4$                 & $44.2 \pm 1.1$ \\
(1D) BiRNN \cite{birnn}&   & -      & 74.2                      & -     & -\\
(1D) BiLSTM \cite{bi_lstm}&   & -& 75.1   & -     & - \\
(1D) Transformer \cite{review}&   &-                         & 78.6     & -     & -\\
(1D) ResNet \cite{res} &   & $ 77.0 \pm 1.5$          & $74.3 \pm 1.7$          & $75.6 \pm 1.6$         & $68.5\pm 1.9$\\
(1D) ResNet \cite{res} + Aug &  \multirow{-7}{*}{\begin{tabular}[c]{@{}c@{}}Cross-Entropy\end{tabular}} &
$45.3 \pm 0.2$ & $37.0 \pm 0.6$  & $45.6 \pm 0.2$    & $37.2 \pm 0.5$ \\ \hline
\multicolumn{1}{|l|}{(1D) Auto Encoder \cite{Soaro}} &
   &
  \multicolumn{1}{c}{$\textbf{69.1} \pm 0.9$} &
  \multicolumn{1}{c|}{$57.4 \pm 1.2$ } &
  \multicolumn{1}{c}{$\textbf{69.2} \pm 1.1$} &
  \multicolumn{1}{c|}{$55.6 \pm 1.3$} \\
(1D) DNN \cite{1DNN}&      &$ \textbf{77.0} \pm 0.6$      & $\textbf{62.8} \pm 0.9$   & $\textbf{72.5} \pm 0.9$      & $\textbf{55.8} \pm 1.1$\\
(1D) ResNet \cite{res} &  & $\textbf{77.8} \pm 0.6$      & $67.0 \pm 1.0$    
& $\textbf{80.7} \pm 0.7$      &  $\textbf{70.1} \pm 0.8$\\
\multicolumn{1}{|l|}{(1D) ResNet \cite{res} + Aug} &
  \multirow{-4}{*}{\begin{tabular}[c]{@{}c@{}}IWL (Ours)\end{tabular}} & $\textbf{45.8} \pm 0.2$    & $\textbf{37.5} \pm 0.5$     & $\textbf{45.9} \pm 0.2$   & $\textbf{37.4} \pm 0.5$               \\ \hline
\multicolumn{1}{|l|}{\multirow{5}{*}{\begin{tabular}[l]{@{}l@{}} (2D) EfficientNetB0 \cite{effi} \\ +  CME (Ours) \end{tabular}}} & CB Focal \cite{clsballoss}  ($\gamma=2$)                   & $24.2 \pm 0.2$                      & $8.1 \pm 0.1$                      & $24.0 \pm 0.3$                 & $10.1 \pm 0.1$        \\ 
\multicolumn{1}{|l|}{} & Class-balanced \cite{clsballoss}   & $80.8 \pm 0.1$                      & $74.9 \pm 0.2$       & $80.7 \pm 0.3$                 & $74.0 \pm 0.4$               \\ 
\multicolumn{1}{|c|}{} & Focal \cite{focal} ($\gamma=2$)                   & $81.8 \pm 0.3$                      & $73.2 \pm 0.6$                      & $81.2 \pm 0.2$                 & $71.5 \pm 0.3$               \\
\multicolumn{1}{|c|}{} & LDAM \cite{labloss} ($\mu = 0.2$, $s = 20$)                 & $79.9 \pm 0.9$                      & $73.8 \pm 0.6$                      & $78.7 \pm 0.7$                 & $72.1 \pm 0.5$               \\
\multicolumn{1}{|c|}{} & Cross-Entropy   & $78.1 \pm 0.7$                      & $69.4 \pm 1.0$                      & $77.5 \pm 0.8$                 & $70.3 \pm 1.1$               \\ 
\hline
(2D) RegNet \cite{reg} + CME&
  \multicolumn{1}{l|}{} & $77.5 \pm 1.1 $    & $71.6 \pm 1.9$     & $80.0 \pm 0.8$          & $71.1 \pm 1.1$ \\
(2D) ResNeXt \cite{res} + CME& \multicolumn{1}{l|}{}   &$78.2 \pm 0.7$                      & $71.2 \pm 0.8$                      & $79.1 \pm 0.4$                 & $70.9 \pm 0.9$\\
(2D) EfficientNetV2 \cite{v2} + CME& \multicolumn{1}{l|}{\multirow{-3}{*}{IWL (Ours)}} &$81.9 \pm 0.5$                      & $72.5 \pm 0.6$                      & $81.7 \pm 0.5$                 & $72.9 \pm 0.8$ \\ \hline
\multicolumn{2}{|l|}{\multirow{2}{*}{\shortstack (2D) EfficientNetB0 \cite{effi} + CME + IWL loss}}    & \cellcolor[HTML]{FFCCC9}$\textbf{83.2} \pm 0.6$ & \cellcolor[HTML]{FFFE65}$\textbf{79.1} \pm 0.8$    & \cellcolor[HTML]{FFCCC9}$\textbf{85.2} \pm 0.7$   & \cellcolor[HTML]{FFFE65}$\textbf{81.0} \pm 0.8$\\     \multicolumn{2}{|l|}{} &                     \cellcolor[HTML]{FFCCC9}$\uparrow 1.3$ & \cellcolor[HTML]{FFFE65}$\uparrow 4.2$    & \cellcolor[HTML]{FFCCC9}$\uparrow 3.5$   & \cellcolor[HTML]{FFFE65}$\uparrow 8.1$\\
\hline
\end{tabular}}
\label{table:compare}
\end{table}
\subsection{Ablation Test}
\subsubsection{Configurations of $\beta$.}\label{secabl1}
The $\beta$ value of the IWL loss function is examined to obtain the optimal one for the model performance. Table \ref{table:beta} shows the results of the experiment carried out with two models: ResNet and ICWA-SE, each involves the IWL loss function with 10 different values of $\beta \in \{0.1, 0.3, 0.5, 0.7, 0.9, 1, 2, 3, 4, 5\}$. It can be seen that the results obtained with different $\beta$ values fluctuated, however, the smaller $\beta$ values give better performances. For the ResNet model, $\beta = 0.1$ gives better accuracy than all other results with $\alpha = 0.5$, and $\beta = 0.3$ gives the best accuracy with the data whose imbalance factor $\alpha = 0.1$.
For ICWA-SE, the best result in both $\alpha$ values was acquired when $\beta=0.3$, both in accuracy and F1-score. The higher $\beta$ values do not improve the performance of the two models, therefore, $\beta = 0.3$ has been used when implementing IWL to other methods for comparison.  

\begin{table}[!h]
\centering
\caption{Quantitative results obtain from experiments with different $\beta$ values.}
\resizebox{\textwidth}{!}{%
\begin{tabular}{|c|cccc|cccc|}
\hline
\multirow{3}{*}{\; $\beta$ \;} & \multicolumn{4}{c|}{$\alpha = 0.05$ (Original)}  & \multicolumn{4}{c|}{$\alpha = 0.01$}  \\ \cline{2-9} 
                      & \multicolumn{2}{c|}{ResNet}                                   & \multicolumn{2}{c|}{CME}             & \multicolumn{2}{c|}{ResNet}                                   & \multicolumn{2}{c|}{CME}             \\ \cline{2-9} 
                      & \multicolumn{1}{c}{\; Accuracy \; } & \multicolumn{1}{c|}{\; F1-score \; } & \multicolumn{1}{c}{\; Accuracy \; } & \;F1-score\; & \multicolumn{1}{c}{\; Accuracy \; } & \multicolumn{1}{c|}{\; F1-score \; } & \multicolumn{1}{c}{\; Accuracy \; } & \; F1-score \;  \\ \hline
0.1          & \cellcolor[HTML]{96FFFB}$80.5 \pm 0.5$   & $72.4\pm 0.5 $   & $81.6 \pm 0.5$     & $73.4 \pm 0.5$       & $79.8 \pm 0.9$   & $70.4 \pm 0.9$     & $82.5 \pm 0.6$    & $75.0 \pm 0.6$      \\
0.3     & $77.8\pm 0.6$    & $67.0 \pm 1.0$      & \cellcolor[HTML]{96FFFB}$83.2 \pm 0.6$     & \cellcolor[HTML]{88E713}$79.1 \pm 0.8$   & \cellcolor[HTML]{96FFFB} $80.7 \pm 0.7$  & $70.1 \pm 0.8$   & \cellcolor[HTML]{96FFFB}$85.2 \pm 0.7$     & \cellcolor[HTML]{88E713} $81.0 \pm 0.8$   \\
0.5     & $76.0 \pm 0.6$     & $65.5 \pm 0.6$  & $82.5 \pm 0.7$      & $78.1 \pm 0.8$   & $77.7\pm 1.0$  & $70.8\pm 1.0$ & $83.6 \pm 1.0$   & $77.0 \pm 0.7$      \\
0.7       & $77.6 \pm 0.7$    & $71.2 \pm 0.9$  & $81.9 \pm 0.5$   & $73.5 \pm 0.5$  & $79.4 \pm 0.6$   & $66.3 \pm 0.5 $   & $83.0 \pm 0.6$   & $75.2 \pm 0.5$      \\
0.9    & $80.1 \pm 0.5$   & $71.3 \pm 0.7$   & $83.4 \pm 0.7$    & $75.0 \pm 0.6$        & $77.3 \pm 0.6$   & \cellcolor[HTML]{88E713} $73.3 \pm 0.6$  & $83.5 \pm 0.7$  & $79.2 \pm 0.7$   \\ 
1     & $79.4 \pm 0.8$     & \cellcolor[HTML]{88E713} $74.5 \pm 0.9$    & $82.9 \pm 0.8$       & $76.0 \pm 0.8$    & $78.7 \pm 1.0$  & $69.9 \pm 0.9$    & $83.2 \pm 0.6$  & $78.5 \pm 0.8$    \\
2     & $78.1 \pm 0.8$  & $71.0\pm 0.9$        & $83.1 \pm 1.1$     & $77.3 \pm 0.9$    & $77.6 \pm 1.0$     & $71.9 \pm 1.1$      & $82.8\pm 0.9$     & $74.9 \pm 1.1$ \\
3     & $78.6 \pm 1.0$    & $71.3 \pm 1.1$  & $82.4 \pm 1.1$   & $75.0 \pm 0.8$   & $78.9 \pm 0.9$   & $70.6 \pm 1.0$    & $80.9 \pm 1.0$    & $77.2 \pm 1.0$     \\
4    & $75.0  \pm 1.2$   & $60.2 \pm 1.2$     & $81.0 \pm 1.0$     & $74.6 \pm 1.1$ & $74.6 \pm 1.2$    & $64.2 \pm 1.1$    & $79.8 \pm 1.0$     & $76.4 \pm 1.1$      \\
5    & $76.6 \pm 1.3$    & $61.8 \pm 1.3$      & $81.8 \pm 1.4$   & $75.7 \pm 1.3$          & $72.3 \pm 1.5$   & $57.5\pm 1.2$     & $80.7 \pm 0.9$    & $76.0 \pm 0.9$      \\
\hline
\end{tabular}}
\label{table:beta}
\end{table}
\subsubsection{Adaptability to Model Architecture Design.} To test the ability to adapt to different models of the proposed CME and IWL loss ($\beta = 0.3$), four models which are RegNet \cite{reg}, ResNeXt \cite{next}, EfficientNetB0 \cite{effi} and EfficientNetV2 \cite{v2} were brought to discussion. The results are presented in Table \ref{table:compare}. As easily seen, these models give higher accuracy than all SOTA models, which is a certain proof of the adaptation of the CME method and IWL loss. The proposed method has the best performance when the EfficientNetB0 model architecture is applied.
\section{Conclusion}
In conclusion, there are some complications inhibiting the performance in ECG classification task of DL models, in particular, long-tailed class distribution, biases and multi-channel temporal data. 
Current endeavors primarily emphasize the enhancement of DL architectures, whereas our proposed method offers a straightforward yet potent approach to directly confront these challenges.
Utilizing the proposed CME technique on channel-wise signals effectively deals with dimensional diversity by minimizing redundant information and improving the focus on data features.
Moreover, we introduce IWL loss to effectively mitigates imbalances within ECG datasets, resulting in improved classification accuracy.
When applying IWL loss, the accuracy of state-of-the-art models (SOTA) increase up to $5\%$ in the CPSC2018 dataset. CME in combination with IWL also surpasses the classification results of other baselines models from $5\%$ to $10\%$. 
%
%
\bibliographystyle{splncs04}
\bibliography{ref}

\end{document}